\documentclass[letterpaper]{article} 
\usepackage{aaai25}  
\usepackage{times}  
\usepackage{helvet}  
\usepackage{courier}  
\usepackage[hyphens]{url}  
\usepackage{graphicx} 
\usepackage{natbib}  
\usepackage{caption} 
\usepackage{algorithm}
\usepackage{algorithmic}

\usepackage{amsmath}
\usepackage{amssymb}    
\usepackage{amsfonts}   
\usepackage{soul}
\usepackage{booktabs}
\usepackage{todonotes}
\usepackage{subcaption}

\usepackage{newfloat}
\usepackage{listings}
\DeclareCaptionStyle{ruled}{labelfont=normalfont,labelsep=colon,strut=off} 
\floatstyle{ruled}
\newfloat{listing}{tb}{lst}{}
\floatname{listing}{Listing}
\title{MABR: Multilayer Adversarial Bias Removal Without Prior Bias Knowledge}
\author{
    Maxwell J. Yin, Boyu Wang, Charles Ling\thanks{Corresponding author}
}
\affiliations{
    Western University\\
    jyin97@uwo.ca, bwang@csd.uwo.ca, charles.ling@uwo.ca
}

\usepackage{bibentry}

\begin{document}

\maketitle

\begin{abstract}
    Models trained on real-world data often mirror and exacerbate existing social biases. Traditional methods for mitigating these biases typically require prior knowledge of the specific biases to be addressed, and the social groups associated with each instance. In this paper, we introduce a novel adversarial training strategy that operates withour relying on prior bias-type knowledge (e.g., gender or racial bias) and protected attribute labels. Our approach dynamically identifies biases during model training by utilizing auxiliary bias detector. These detected biases are simultaneously mitigated through adversarial training. Crucially, we implement these bias detectors at various levels of the feature maps of the main model, enabling the detection of a broader and more nuanced range of bias features. Through experiments on racial and gender biases in sentiment and occupation classification tasks, our method effectively reduces social biases without the need for demographic annotations. Moreover, our approach not only matches but often surpasses the efficacy of methods that require detailed demographic insights, marking a significant advancement in bias mitigation techniques.
\end{abstract}
    
%
\begin{links}
    \link{Code}{https://github.com/maxwellyin/MABR}
\end{links}

\section{Introduction}

Neural natural language processing (NLP) models are known to exhibit social biases, with protected attributes like gender and race serving as confounding variables \citep{blodgett-etal-2020-language,bansal2022survey}. These attributes can create spurious correlations with task response variables, leading to biased predictions. This issue manifests across various NLP tasks, such as machine translation \citep{cho-etal-2019-measuring, stanovsky-etal-2019-evaluating}, dialogue generation \citep{liu-etal-2020-gender}, and sentiment analysis \citep{kiritchenko-mohammad-2018-examining}.

Adversarial approaches are widely used to reduce bias related to protected attributes. In these methods, the text encoder strives to obscure protected attributes so that the discriminator cannot identify them \citep{li-etal-2018-towards, zhang2018mitigating, han-etal-2021-diverse}. However, these methods require training examples labeled with protected attributes, which presents several challenges. First, we may not be aware of the specific biases, like gender or age bias, that require mitigation \citep{orgad-belinkov-2023-blind}. Second, obtaining protected labels can be difficult due to privacy regulations and ethical concerns, leading to few users publicly disclosing their protected attributes \citep{han-etal-2021-decoupling}. Moreover, prior research has typically focused on mitigating a single type of bias \citep{schuster-etal-2019-towards, clark-etal-2019-dont, utama-etal-2020-mind}. However, in practice, corpora often contain multiple types of biases, each with varying levels of detection difficulty.

In this paper, we address the challenge of bias removal without prior knowledge of bias labels by proposing a Multilayer Adversarial Bias Removal (MABR) framework. We introduce a series of auxiliary classifiers as bias detectors. The rationale behind using multiple classifiers is to capture different aspects and levels of bias present in the data. Each classifier operates on different layers of the main model's encoder, based on the insight that different layers of the encoder may capture different aspects of bias. Lower-level feature maps may capture word-level biases, such as associating words like ``nurse" or ``secretary" predominantly with female pronouns or contexts, and words like ``engineer" or ``pilot" with male pronouns. Higher-level feature maps may capture more subtle gender biases, such as associating leadership qualities with male-associated terms or nurturing qualities with female-associated terms, or inferring competence and ambition based on gendered names or contexts. These biases manifest in more nuanced ways, such as assuming managerial roles are more suited to one gender over another, reflecting societal stereotypes in professional settings.

Once biased samples are detected, we apply adversarial training to mitigate these biases. We introduce domain discriminators at each layer of the main model's encoder. The goal of the adversarial training is to make the representations learned by the main model invariant to the biases identified by the auxiliary classifiers. To achieve this, we employ a Reverse Gradient Layer during backpropagation, which ensures that the main model generates feature representations that are indistinguishable with respect to the domain discriminators. This process encourages the alignment of feature distributions between biased and unbiased samples, thereby reducing the influence of biased features on the model's predictions.

However, this approach alone is insufficient. The bias detector tends to detect relatively easy samples where the biased features are obvious or the sentence structure is simple. Building on the findings of \citet{liu2021just}, we recognize that standard training of language models often results in models with low average test errors but high errors on specific groups of examples. These performance disparities are particularly pronounced when spurious correlations are present. Therefore, we also consider training examples misclassified by the main model as hard biased samples, supplementing the samples detected by the bias detector.

We conduct experiments on two English NLP tasks and two types of social demographics: sentiment analysis with gender and occupation classification with race. Our MABR method successfully reduces bias, sometimes even outperforming methods that use demographic information. This indicates that MABR may offer a more robust solution for bias mitigation compared to other existing methods.

Our contributions are as follows:
\begin{enumerate}
    \item We introduce MABR, an adversarial bias removal method that does not require prior knowledge of specific biases.
    \item We enhance bias detection in MABR by enabling it on all layers of the main model’s encoder, capturing various types of biases.
    \item We demonstrate that MABR can successfully reduce bias without protected-label data and is robust across different tasks and datasets.
\end{enumerate}

\section{Related Work}

Research suggests various methods for mitigating social biases in NLP models applied to downstream tasks. Some approaches focus on preprocessing the training data, such as converting biased words to neutral alternatives \citep{de2019bias} or to those that counteract bias \citep{zhao-etal-2018-gender}, or balancing each demographic group in training \citep{zhao-etal-2018-gender, wang2019balanced, lahoti2020fairness, han-etal-2022-balancing}. Others focus on removing demographic information from learned representations, for instance, by applying post-hoc methods to the neural representations of a trained model \citep{ravfogel-etal-2020-null, ravfogel2022linear, iskander-etal-2023-shielded}. Adversarial training is also a common strategy \citep{li-etal-2018-towards, zhang2018mitigating, elazar-goldberg-2018-adversarial, wang2019balanced, han-etal-2021-diverse}. However, all these methods require prior knowledge of the specific bias to be addressed, such as gender bias. Furthermore, many of these approaches depend on demographic annotations for each data instance. For example, to address gender bias, each data sample must be annotated to indicate whether it pertains to a male or female subject. In contrast, our method does not require any prior knowledge about the bias. Additionally, while the authors of these studies select hyperparameters based on the fairness metrics they aim to optimize, we choose our hyperparameters without explicitly measuring fairness metrics.

Recent studies have also explored fairness in machine learning through alternative approaches, such as discovering intersectional unfairness \citep{xu2024intersectional}, learning fairness across multiple subgroups \citep{shui2022learning}, and aligning representations implicitly for fair learning \citep{shui2022fair}. Other work in related areas has proposed leveraging prompt-based learning \citep{yin-etal-2024-source} or masking mechanisms \citep{10.1162/tacl_a_00669} to mitigate domain gaps. These methods contribute to advancing fairness research but still differ from our approach, which avoids both demographic annotations and prior bias knowledge.

\begin{figure*}
  \centering
  \includegraphics[width=\textwidth]{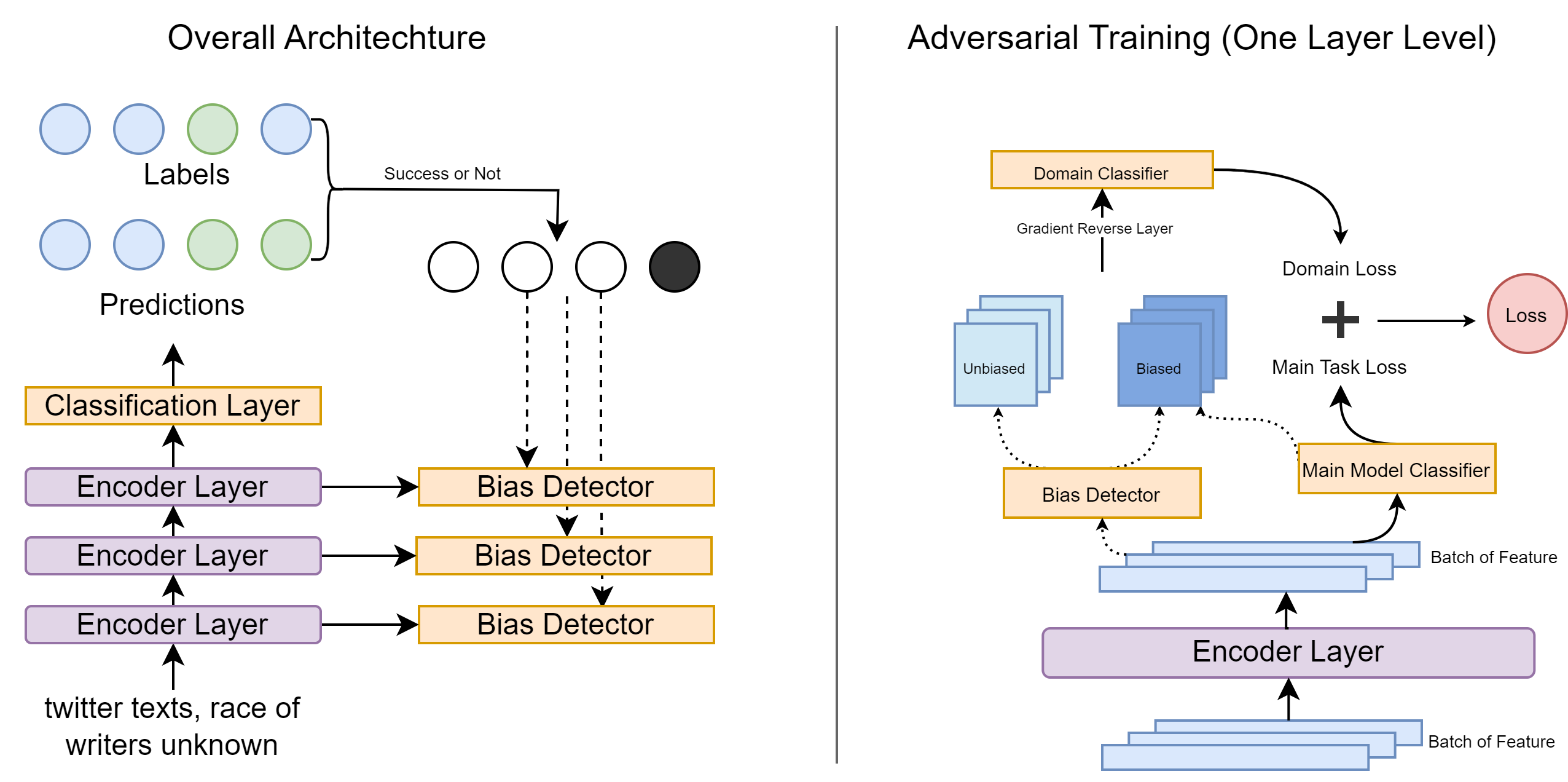}
  \caption{Schematic Overview of the MABR Framework. The left panel illustrates the overall architecture of the model for main task and bias detection. The right panel details the domain adversarial training process upon each encoder layer.}
  \label{fig:model}
\end{figure*}

\section{Methodology}
\subsection{Problem Formulation}

We consider the problem of general multi-class classification. The dataset $\mathcal{D} = \{(x_i, y_i, z_i)\}_{i=1}^N$ comprises triples consisting of an input $x_i \in \mathcal{X}$, a label $y_i \in \mathcal{Y}$, and a protected attribute $z_i \in \mathcal{Z}$, which corresponds to a demographic group, such as gender. The attribute $z_i$ is unknown, meaning it is not accessible during training stages. Our objective is to learn a mapping $f_M: \mathcal{X} \rightarrow \mathbb{R}^{|\mathcal{Y}|}$, where $f_M$, referred to as the main model, is resilient to demographic variations introduced by $z_i$, with $|\mathcal{Y}|$ denoting the number of classes.

The model's fairness is evaluated using various metrics. A fairness metric maps a model's predictions and the associated protected attributes to a numerical measure of bias: $M: (\mathbb{R}^{|\mathcal{Y}|}, \mathcal{Z}) \rightarrow \mathbb{R}$. The closer the absolute value of this measure is to 0, the fairer the model is considered to be.

\subsection{Bias Detection}

Since the protected attribute $\mathcal{Z}$ is unknown, we detect possible biased samples automatically and dynamically. To achieve this, we introduce a bias detector for each layer of the encoder, as depicted in Fig. \ref{fig:model}. Given the embedding output of a specific layer, the bias detector on that layer is trained to predict whether the main model will successfully predict the correct label for the main task for each training sample. Let \( L \) denote the total number of layers in the encoder. It is formulated as $f_{B_l} : g_l(\mathcal{X}) \rightarrow \mathbb{R}^{|s_l|}$ for each layer $l$,  where \( g_l(x) \) represents the output embedding of the \( l_{th}\) layer and \( l \) ranges from 1 to \( L \). Here, $s_l$ is an indicator function defined as: $s_l = \mathbb{I}(f_M(x) = y)$, which is dynamic and changes across different epochs of the training process. Notably, the bias detector has no knowledge of the original task, and the prediction is made without access to the main task label. The intuition behind this approach is that if the bias detector can successfully predict the main model's behavior based solely on a single embedding layer output, without access to the task label, it indicates that the main model likely relies on a specific bias feature as a shortcut, leading to shallow decision-making.

Initially, we train both the main model and the bias detectors using the standard training process, where both models are optimized using cross-entropy loss.

The cross-entropy loss for the main model, represented as $\mathcal{L}_{\text{main}}$, is defined in the equation below:
\begin{equation}
\mathcal{L}_\text{main} = -\frac{1}{N} \sum_{i=1}^{N} \sum_{c=1}^{|\mathcal{Y}|} y_{i,c} \log \left( f_{M,c}(x_i) \right)
\end{equation}
where $N$ is the number of training samples,  $y_{i,c}$ is a binary indicator (0 or 1) indicating whether class label $c$ is the correct classification for sample $i$, and $f_{M,c}(x_i)$ is the predicted probability of the main model for class $c$.

The cross-entropy loss for the bias detector at the $l_{th}$ layer,$\mathcal{L}_\text{bias}$, is defined as follows:
\begin{align}
\mathcal{L}_\text{bias}^l &= -\frac{1}{N} \sum_{i=1}^{N} \left( s_i^l \log \left( f_{B_l}(g_l(x_i)) \right) \right. \nonumber \\
& \quad + \left. (1 - s_i^l) \log \left( 1 - f_{B_l}(g_l(x_i)) \right) \right)
\end{align}
where \( s_i^l = \mathbb{I}(f_M(x_i) = y_i) \) is an indicator function that denotes whether the main model's prediction is correct for sample $i$ at layer $l$, and $f_{B_l}(g_l(x_i))$ is the predicted probability of the bias detector at layer $l$.

The total loss for the bias detectors across all layers, \(\mathcal{L}_\text{bias}\), is obtained by summing the losses from each layer, as formulated below:
\begin{equation}
    \mathcal{L}_\text{bias} = \sum_{l=1}^{L} \mathcal{L}_\text{bias}^l
\end{equation}

After the initial training phase, we utilize the bias detectors \(\mathcal{B} = \{ f_{B_l} \}_{l=1}^L\) to identify biased samples. If the bias detector can predict whether the main model is correct or incorrect on a sample (i.e., \(\sigma(f_{B}(x))\) is high), without knowing the task at hand, then the sample likely contains some simple but biased features. This intuition aligns with the claim that in the context of complex language understanding tasks, all simple feature correlations are spurious \cite{gardner-etal-2021-competency,orgad-belinkov-2023-blind}. Therefore, the samples for which the bias detector predicts a score higher than the threshold \(\tau\) are considered biased samples, where \(\tau\) is a hyperparameter.

Nevertheless, this approach tends to detect samples with more apparent biases or simpler sentence structures. To address this limitation, we incorporate insights from \citet{liu2021just}, which highlight that language models trained with standard methods can achieve low average test errors while exhibiting high errors on certain groups due to spurious correlations. Consequently, we also consider misclassified training examples as hard biased samples. Formally, for a sample \(x_i\), if \(\hat{y}_i = f_M(x_i) \neq y_i\), it is deemed a hard biased sample. This supplementary set of hard biased samples enhances the identification of biased instances beyond those detected by the bias detector alone.

\subsection{Adversarial Training}

As illustrated in the right part of Fig. \ref{fig:model}, we employ an adversarial training process to mitigate the biases identified by the bias detectors and hard biased samples. This process involves two primary components: the main model \( f_M \) and a set of domain discriminators \( \mathcal{G} = \{ G_l \}_{l=1}^L \). The goal of adversarial training is to make the representations learned by the main model invariant to the identified biases.

The main model \( f_M \) can be decomposed into an encoder \( g \) and a classifier \( h_M \), such that \( f_M = h_M \circ g \). Each domain discriminator \( G_l \) attempts to predict whether a sample is biased or not based on the representations generated by \( g_l \).

For adversarial training, we employ the Reverse Gradient Layer \citep{ganin2014unsupervised} to ensure that the main model learns to generate representations that are invariant to the identified biases. The Reverse Gradient Layer functions by reversing the gradient during backpropagation, thereby encouraging the main model to produce feature representations that are indistinguishable with respect to the domain discriminators. 

The adversarial training is conducted at each layer of the encoder separately. The adversarial loss for a sample \(x_i\) at layer \(l\) is computed as follows:
\begin{align}
\mathcal{L}_\text{adv}^l &= -\frac{1}{N} \sum_{i=1}^{N} \left( z_i^l \log \left( G_l(g_l(x_i)) \right) \right. \nonumber \\
& \quad + \left. (1 - z_i^l) \log \left( 1 - G_l(g_l(x_i)) \right) \right)
\end{align}
where \( z_i^l \) is an indicator variable that denotes whether the sample \( x_i \) is considered biased (i.e., identified by the bias detector or misclassified by the main model).

The total loss for the adversarial training is a combination of the main model's cross-entropy loss, the bias detector's cross-entropy loss, and the adversarial loss at each layer. The combined loss function is given by:
\begin{equation}
\mathcal{L}_\text{total} = \mathcal{L}_\text{main} + \sum_{l=1}^{L} \mathcal{L}_\text{bias}^l + \sum_{l=1}^{L} \mathcal{L}_\text{adv}^l
\end{equation}

During backpropagation, the weights of the encoder \( g \) are updated to minimize the total loss \(\mathcal{L}_\text{total}\). Let \( \theta_g \) represent the weights of the encoder \( g \). The update rule for \( \theta_g \) is:
\begin{equation}
    \theta_g \leftarrow \theta_g - \eta \left( \frac{\partial \mathcal{L}_\text{main}}{\partial \theta_g} - \sum_{l=1}^{L} \frac{\partial \mathcal{L}_\text{adv}^l}{\partial \theta_g} \right)
\end{equation}

It is important to note that the gradient contribution from the adversarial loss \(\mathcal{L}_\text{adv}^l\) is reversed by the Reverse Gradient Layer, and the gradient from the bias detectors is not used for updating \( \theta_g \). The whole procedure is detailed in Algorithm~\ref{alg:adversarial_training}.

\begin{algorithm}[t]
\caption{Adversarial Training with Bias Detection and Mitigation}
\label{alg:adversarial_training}
\begin{algorithmic}[1]
\REQUIRE Dataset $\mathcal{D} = \{(x_i, y_i, z_i)\}_{i=1}^N$
\REQUIRE Encoder $g$, Classifier $h_M$, Bias Detectors $\mathcal{B} = \{ f_{B_l} \}_{l=1}^L$, Domain Discriminators $\mathcal{D} = \{ D_l \}_{l=1}^L$
\REQUIRE Threshold $\tau$, Learning rate $\eta$

\STATE Initialize the main model $f_M = h_M \circ g$
\STATE Initialize bias detectors $\mathcal{B}$ and domain discriminators $\mathcal{D}$

\STATE \textbf{Phase 1: Initial Training (1 epoch)}
\FOR{each mini-batch $\mathcal{M}$ in $\mathcal{D}$}
    \STATE Compute main model outputs: $f_M(x)$
    \STATE Compute cross-entropy loss: $\mathcal{L}_\text{main}$
    \STATE Update main model parameters to minimize $\mathcal{L}_\text{main}$
    \STATE Compute bias detector outputs: $f_{B_l}(g_l(x))$
    \STATE Compute cross-entropy loss for bias detectors: $\mathcal{L}_\text{bias}^l$
    \STATE Update bias detector parameters to minimize $\mathcal{L}_\text{bias}^l$
\ENDFOR

\STATE \textbf{Phase 2: Adversarial Training (T epochs)}
\FOR{epoch $= 1$ to $T$}
    \FOR{each mini-batch $\mathcal{M}$ in $\mathcal{D}$}
        \STATE Compute main model outputs: $f_M(x)$
        \STATE Compute bias detector outputs: $f_{B_l}(g_l(x))$
        \STATE Identify biased samples using threshold $\tau$ and misclassified main model samples
        \STATE Compute adversarial loss for domain discriminators: $\mathcal{L}_\text{adv}^l$
        \STATE Compute total loss: $\mathcal{L}_\text{total} = \mathcal{L}_\text{main} + \sum_{l=1}^{L} \mathcal{L}_\text{bias}^l + \sum_{l=1}^{L} \mathcal{L}_\text{adv}^l$
        \STATE Update encoder parameters to minimize $\mathcal{L}_\text{total}$ with reversed gradient for $\mathcal{L}_\text{adv}^l$
    \ENDFOR
\ENDFOR

\STATE \textbf{Output:} Trained main model $f_M$, bias detectors $\mathcal{B}$, and domain discriminators $\mathcal{D}$

\end{algorithmic}
\end{algorithm}

\section{Experiments}

\subsection{Tasks and Models}

In our experiments, we investigate two classification tasks, each associated with a distinct type of bias:

\subsubsection{Sentiment Analysis and Race}

Following the methodology of previous research \citep{elazar-goldberg-2018-adversarial,orgad-belinkov-2023-blind}, we employ a dataset from \citet{blodgett-etal-2016-demographic} that consists of 100,000 tweets to explore dialect differences in social media language. This dataset allows us to analyze racial identity by categorizing each tweet as either African American English (AAE) or Mainstream US English (MUSE), commonly referred to as Standard American English (SAE). The classification leverages the geographical information of the tweet authors. Additionally, \citet{elazar-goldberg-2018-adversarial} used emojis embedded in tweets as sentiment indicators to facilitate the sentiment classification task.

\subsubsection{Occupation Classification and Gender Bias}

Following previous research \citep{orgad-belinkov-2023-blind}, we utilize the dataset provided by \citet{de2019bias}, which comprises 400,000 online biographies, to examine gender bias in occupational classification. The task involves predicting an individual's occupation using a portion of their biography, specifically excluding the first sentence that explicitly mentions the occupation. The protected attribute in this context is gender, and each biography is labeled with binary gender categories based on the pronouns used within the text, reflecting the individual's self-identified gender.

\subsection{Metrics}

Research by \citet{orgad-belinkov-2022-choose} demonstrates that different fairness metrics can respond variably to debiasing methods. Specifically, methods designed to improve fairness according to one metric may actually worsen outcomes when measured by another. Therefore, to achieve a comprehensive analysis of the performance of our method and previous baselines, we measure multiple metrics.

\subsubsection{True Positive Rate gap}

The True Positive Rate (TPR) gap indicates the difference in performance between two demographic groups, such as females versus males. For gender, we measure the TPR gap for label \( y \) as \( GAP_{TPR,y} = |TPR^F_y - TPR^M_y| \). To provide a more comprehensive assessment, we calculate the root-mean-square form of the TPR gap (denoted \( TPR_{RMS} \)), which is \( \sqrt{\frac{1}{|Y|} \sum_{y \in Y} (GAP_{TPR,y})^2} \), following previous research \citep{ravfogel-etal-2020-null,ravfogel2022linear,orgad-belinkov-2023-blind}.

\subsubsection{Independence}

This metric evaluates the statistical independence between the model's predictions and the protected attributes. According to the independence rule (demographic parity), the probability of a positive prediction should be the same regardless of the protected attribute. To measure this, we calculate the Kullback-Leibler (KL) divergence between two distributions: \( KL(P(\hat{Y}), P(\hat{Y} | Z = z)) \), \( \forall z \in \mathcal{Z} \). Summing these values over \( z \) gives a single measure reflecting the model's independence. This metric does not consider the true labels (gold labels); instead, it intuitively measures how much the model's behavior varies across different demographic groups.

\subsection{Sufficiency}

This metric measures the statistical dependence between the target label given the model's prediction and the protected attributes. It uses the Kullback-Leibler divergence between two distributions: \( KL(P(y|r), P(y|r, z = z)) \), for all \( r \in \mathcal{Y} \) and \( z \in \mathcal{Z} \). The values are summed over \( r \) and \( z \) to produce a single measure. Related to calibration and precision gap, this metric assesses if a model disproportionately favors or penalizes a specific demographic group \citep{liu2019implicit}.

\subsection{Implementation Details}

We experiment with BERT \citep{devlin2018bert} and DeBERTa-v1 \citep{he2020deberta} as backbone models, utilizing the transformer model as a text encoder with its output fed into a linear classifier. The text encoder and linear layer are fine-tuned for the downstream task. We implement the MABR framework using the Huggingface Transformers library \citep{wolf-etal-2020-transformers}. The batch size is set to 64, enabling dynamic adversarial training per batch. We set the learning rate to 1e-3 for the bias detector and domain classifier, and 2e-5 for the model. The threshold \(\tau\) is selected to ensure approximately 30\% of samples fall outside it after initial training. For training epochs, we balance task accuracy and fairness using the ``distance to optimum'' (DTO) criterion introduced by \citet{han-etal-2022-balancing}. Model selection is performed without a validation set with demographic annotations, choosing the largest epoch while limiting accuracy reduction. We use 0.98 of the maximum achieved accuracy on the task as the threshold to stop training. Other hyperparameters follow the default settings provided by the Transformers library.

\subsection{Baselines}

\begin{table*}[ht]
    \centering
    \begin{tabular}{lcccc|cccc}
    \toprule
              & \multicolumn{4}{c}{BERT} & \multicolumn{4}{c}{RoBERTa} \\
    \cmidrule(r){2-5} \cmidrule(l){6-9} 
              & Acc \(\uparrow\) & TPR\textsubscript{RMS} \(\downarrow\) & Indep \(\downarrow\) & \multicolumn{1}{c|}{Suff \(\downarrow\)} & Acc \(\uparrow\) & TPR\textsubscript{RMS} \(\downarrow\) & Indep \(\downarrow\) & Suff \(\downarrow\) \\
    \midrule
    Finetuned & \textbf{0.771} & 0.243 & 0.039 & \multicolumn{1}{c|}{0.028} & \textbf{0.779} & 0.261 & 0.035 & 0.031 \\
    INLP      & 0.753 & 0.198 & 0.021 & \multicolumn{1}{c|}{0.025} & 0.647 & \textbf{0.088} & 0.010 & 0.030 \\
    RLACE     & 0.739 & 0.140 & 0.009 & \multicolumn{1}{c|}{0.021} & 0.751 & 0.157 & 0.014 & 0.032 \\
    JTT       & 0.762 & 0.191 & 0.014 & \multicolumn{1}{c|}{0.028} & 0.753 & 0.185 & 0.013 & \textbf{0.026} \\
    BLIND     & 0.759 & 0.202 & 0.029 & \multicolumn{1}{c|}{0.024} & 0.741 & 0.213 & 0.024 & 0.033 \\
    \midrule
    MABR      & 0.766 & \textbf{0.137} & \textbf{0.003} & \multicolumn{1}{c|}{\textbf{0.021}} & 0.763 & 0.126 & \textbf{0.006} & 0.028 \\
    -multi    & 0.768 & 0.145 & 0.010 & \multicolumn{1}{c|}{0.025} & 0.762 & 0.162 & 0.014 & 0.033 \\
    -hard     & 0.768 & 0.139 & 0.006 & \multicolumn{1}{c|}{0.022} & 0.763 & 0.142 & 0.009 & 0.031 \\
    JTT-Disc + AdvTrain & 0.768 & 0.173 & 0.013 & \multicolumn{1}{c|}{0.027} & 0.763 & 0.169 & 0.017 & 0.034 \\
    Bias-Disc + Upweight & 0.766 & 0.168 & 0.011 & \multicolumn{1}{c|}{0.025} & 0.762 & 0.160 & 0.015 & 0.033 \\
    \bottomrule
    \end{tabular}
    \caption{Performance metrics on the sentiment analysis task, averaged over 5 independent experimental runs.}
    \label{tab:main-results-sentiment}
    \end{table*}

    \begin{table*}[ht]
        \centering
        \begin{tabular}{lcccc|cccc}
        \toprule
                  & \multicolumn{4}{c}{BERT} & \multicolumn{4}{c}{RoBERTa} \\
        \cmidrule(r){2-5} \cmidrule(l){6-9} 
                  & Acc \(\uparrow\) & TPR\textsubscript{RMS} \(\downarrow\) & Indep \(\downarrow\) & \multicolumn{1}{c|}{Suff \(\downarrow\)} & Acc \(\uparrow\) & TPR\textsubscript{RMS} \(\downarrow\) & Indep \(\downarrow\) & Suff \(\downarrow\) \\
        \midrule
        Finetuned & \textbf{0.869} & 0.135 & 0.149 & \multicolumn{1}{c|}{1.559} & \textbf{0.863} & 0.132 & 0.144 & 1.600 \\
        INLP      & 0.857 & 0.131 & 0.137 & \multicolumn{1}{c|}{1.216} & 0.851 & 0.123 & 0.132 & 1.052 \\
        RLACE     & 0.868 & 0.133 & 0.144 & \multicolumn{1}{c|}{1.413} & 0.852 & 0.124 & 0.127 & 1.362 \\
        JTT       & 0.849 & 0.132 & 0.132 & \multicolumn{1}{c|}{1.417} & 0.844 & 0.139 & 0.139 & 1.397 \\
        BLIND     & 0.826 & 0.136 & 0.123 & \multicolumn{1}{c|}{1.097} & 0.839 & 0.123 & 0.122 & 0.906 \\
        \midrule
        MABR      & 0.857 & \textbf{0.101} & \textbf{0.099} & \multicolumn{1}{c|}{\textbf{1.031}} & 0.852 & \textbf{0.109} & \textbf{0.100} & \textbf{0.821} \\
        -multi    & 0.859 & 0.128 & 0.112 & \multicolumn{1}{c|}{1.054} & 0.853 & 0.121 & 0.117 & 0.907 \\
        -hard     & 0.858 & 0.119 & 0.111 & \multicolumn{1}{c|}{1.033} & 0.853 & 0.114 & 0.116 & 0.883 \\
        JTT-Disc + AdvTrain & 0.855 & 0.129 & 0.118 & \multicolumn{1}{c|}{1.061} & 0.850 & 0.133 & 0.125 & 0.876 \\
        Bias-Disc + Upweight & 0.856 & 0.131 & 0.120 & \multicolumn{1}{c|}{1.059} & 0.851 & 0.137 & 0.121 & 0.892 \\
        \bottomrule
        \end{tabular}
        \caption{Performance metrics on the occupation classification task, averaged over 5 independent experimental runs.}
        \label{tab:main-results-occupation}
        \end{table*}

We compare MABR with the following baselines:

\subsubsection{Finetuned}
The MABR model architecture, trained for downstream tasks without any debiasing mechanisms.

\subsubsection{INLP \citep{ravfogel-etal-2020-null}}
A post-hoc method that trains linear classifiers to predict attributes, then projects representations onto their null-space to remove attribute information and mitigate bias.

\subsubsection{R-LACE \citep{ravfogel2022linear}}
Eliminates specific concepts from neural representations using a constrained minimax optimization framework. It employs a projection matrix to remove the linear subspace corresponding to the targeted concept, preventing linear predictors from recovering it.

\subsubsection{BLIND \citep{orgad-belinkov-2023-blind}}
Identifies biased samples through an auxiliary model and reduces their weight during training. Effective for single biases but lacks broader anti-bias capabilities.

\subsubsection{JTT \citep{liu2021just}}
A two-stage framework: first identifies high-loss examples with empirical risk minimization (ERM), then upweights these in the final training to boost worst-group performance without group annotations.

\section{Results}

\subsection{Overall Results}

Tables \ref{tab:main-results-sentiment} and \ref{tab:main-results-occupation} present the performance metrics for various models on the sentiment analysis and occupation classification tasks, respectively. The vanilla fine-tuning baseline yields the highest accuracy but also the worst bias (highest fairness metrics) for both BERT and RoBERTa, and across both tasks. This outcome is expected due to the inherent trade-off between fairness and performance.

\subsubsection{Sentiment Analysis}

For the sentiment analysis task (Table \ref{tab:main-results-sentiment}), MABR effectively reduces bias. On BERT, compared to the finetuned baseline, MABR lowers $TPR_{RMS}$ by 10.6\% and Independence by 3.6\%, with only a 0.5\% drop in accuracy. While R-LACE achieves comparable fairness, its accuracy decreases by 3.2\%. JTT and BLIND show similar accuracy to MABR but fall short in mitigating bias consistently across metrics, with varying performance depending on the metric.

For RoBERTa, MABR also demonstrates strong bias reduction, decreasing $TPR_{RMS}$ by 13.5\% and Independence by 2.9\%, with a 1.6\% drop in accuracy. Although INLP achieves the largest $TPR_{RMS}$ reduction (17.3\%), it suffers a substantial 13.2\% accuracy drop. JTT marginally outperforms MABR in sufficiency reduction, but overall, MABR offers the best balance of fairness and accuracy.

\subsubsection{Occupation Classification}

For the occupation classification task (Table \ref{tab:main-results-occupation}), the finetuned baseline shows less pronounced bias, likely due to the dataset’s lower inherent bias. Nevertheless, MABR significantly improves fairness, reducing $TPR_{RMS}$ by 2.4\% and Independence by 5.0\%, with only a 1.2\% accuracy drop. R-LACE achieves slightly higher accuracy but offers limited bias reduction, with $TPR_{RMS}$ decreasing by only 0.2\%.

For RoBERTa, MABR matches R-LACE in accuracy but provides superior fairness, reducing $TPR_{RMS}$ by 2.3\% and Independence by 4.4\%. MABR consistently outperforms baselines across models and metrics, underscoring its robustness in mitigating bias.

\subsection{Ablation Study}

To better understand our proposed framework, we conducted ablation studies to evaluate the effectiveness of each component. The results are shown in Tables \ref{tab:main-results-sentiment} and \ref{tab:main-results-occupation}. The notation ``-multi'' denotes the removal of the multi-layer bias detection component, and the notation ``-hard'' signifies the omission of the adversarial training with hard examples. Additionally, we analyzed two variants: ``JTT-Disc + AdvTrain,'' which combines JTT's bias discovery with our adversarial training, and ``Bias-Disc + Upweight,'' which integrates our bias detection with JTT's upweighting strategy.

For the sentiment analysis task (Table \ref{tab:main-results-sentiment}), removing the multi-layer bias detection component (``-multi'') results in a slight increase in accuracy but worsens bias performance, with $TPR_{RMS}$ rising by 0.8\% and Independence by 0.7\%. Similarly, omitting the hard example detection process (``-hard'') leads to an increase in bias metrics, with $TPR_{RMS}$ increasing by 0.2\% and Independence by 0.3\%. The ``JTT-Disc + AdvTrain'' variant performs better than JTT alone, as adversarial training mitigates bias more effectively, but its simpler bias discovery mechanism limits its performance. Meanwhile, the ``Bias-Disc + Upweight'' variant improves bias metrics compared to JTT but underperforms MABR, as upweighting at the final layer lacks the nuanced mitigation provided by multi-layer processing. These findings emphasize the importance of both multi-layer bias detection and hard example training, with the former having a more substantial impact.

For the occupation classification task (Table \ref{tab:main-results-occupation}), removing the multi-layer bias detection component (``-multi'') leads to a decrease in accuracy and worsened bias performance, with $TPR_{RMS}$ rising by 2.7\% and Independence by 1.3\%. Similarly, omitting the hard example detection process (``-hard'') increases $TPR_{RMS}$ by 1.8\% and Independence by 1.2\%. The ``JTT-Disc + AdvTrain'' and ``Bias-Disc + Upweight'' variants follow similar trends as in the sentiment analysis task, improving upon JTT in bias mitigation but failing to reach MABR's level of effectiveness due to limitations in either bias detection or mitigation strategies.

Considering the results across both tasks, MABR achieved a larger reduction in bias for sentiment analysis (e.g., a 10.6\% decrease in $TPR_{RMS}$ compared to the finetuning baseline) than for occupation classification (e.g., a 3.4\% decrease in $TPR_{RMS}$). This highlights the critical role of MABR's multi-layer bias detection and adversarial training in addressing deeply ingrained biases, demonstrating their effectiveness in enhancing fairness across tasks with complex or subtle biases.

\section{Layer Level Analysis}

\begin{figure}[t]
    \centering
    \includegraphics[width=\columnwidth]{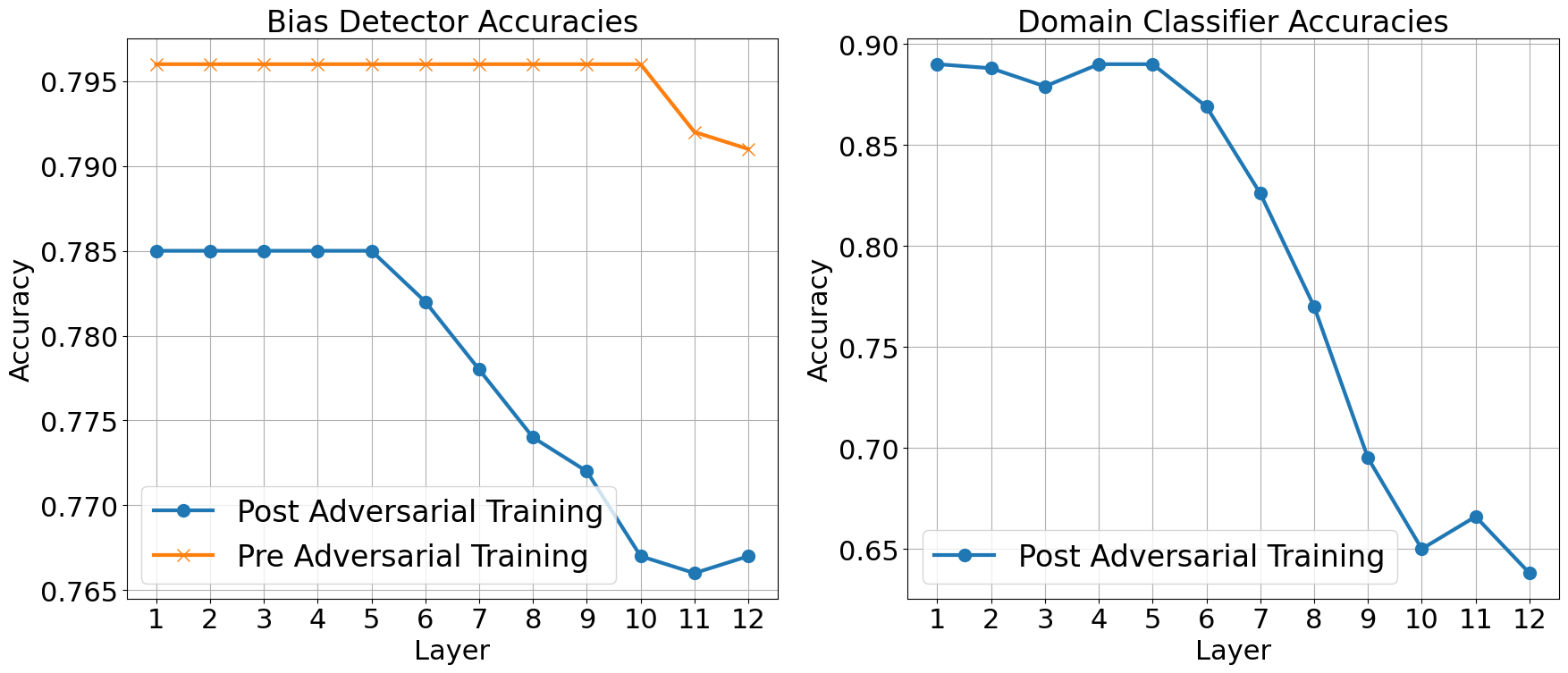}
    \caption{Accuracies for each layer of domain adversarial training components when training with BERT on the sentiment classification task.}
    \label{fig:validation_accuracies}
\end{figure}

\begin{figure}[t]
    \centering
    \includegraphics[width=\columnwidth]{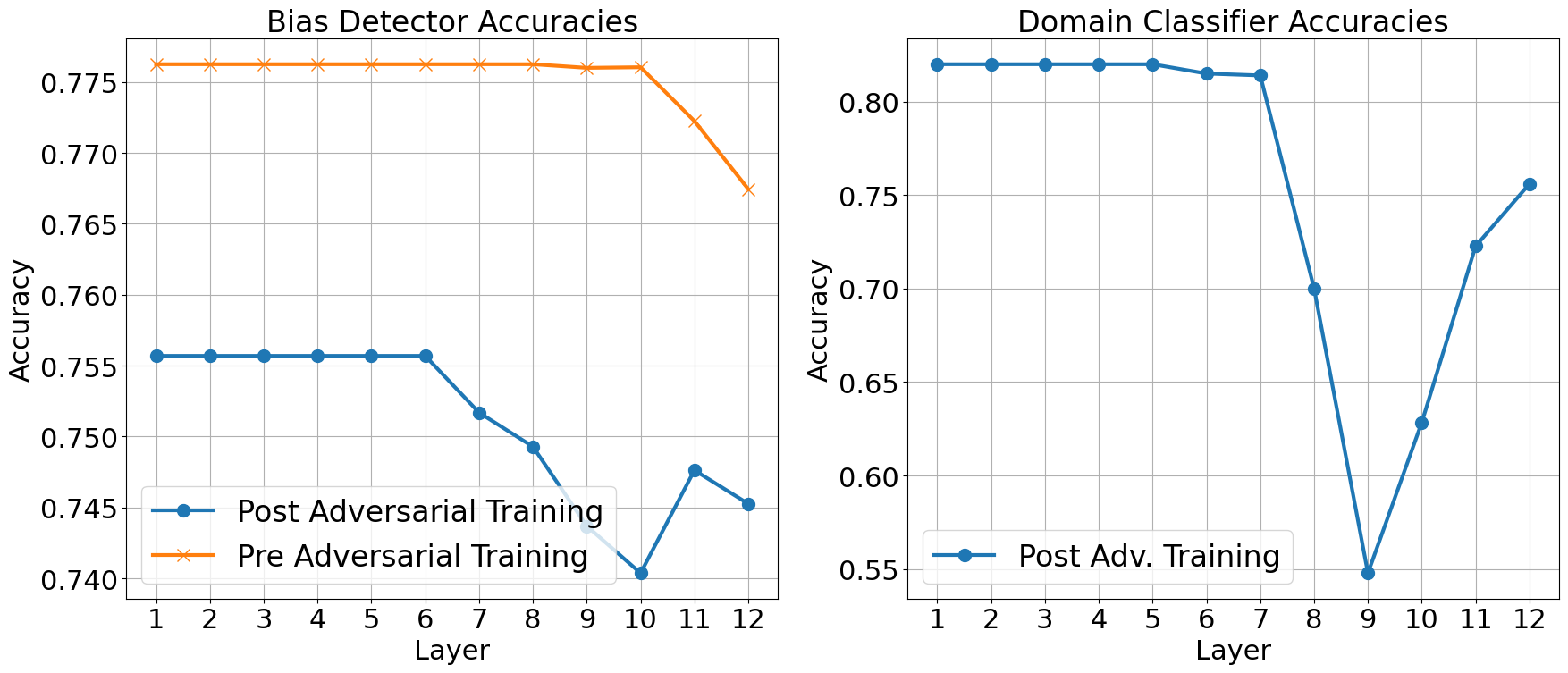}
    \caption{Accuracies for each layer of domain adversarial training components when training with Roberta on the sentiment classification task.}
    \label{fig:validation_accuracies-roberta}
\end{figure}

\begin{table}[t]
\centering
\begin{tabular}{lcccc}
\toprule
                      & Acc \(\uparrow\) & TPR\textsubscript{RMS} \(\downarrow\) & Indep \(\downarrow\) & Suff \(\downarrow\) \\
\midrule
MABR                 & 0.766 & 0.137 & 0.003 & 0.021 \\
- layer[1:5]         & 0.766 & 0.140 & 0.005 & 0.022 \\
- layer[6:9]         & 0.767 & 0.142 & 0.005 & 0.022 \\
- layer[10:12]       & 0.767 & 0.142 & 0.007 & 0.023 \\
\bottomrule
\end{tabular}
\caption{Performance metrics with the bias detector and domain classifier removed at specific layer levels during adversarial training using the MABR method on the sentiment analysis task with BERT.}
\label{tab:bert-remove-specific-layers}
\end{table}

\begin{table}[ht]
\centering
\begin{tabular}{lcccc}
\toprule
                      & Acc \(\uparrow\) & TPR\textsubscript{RMS} \(\downarrow\) & Indep \(\downarrow\) & Suff \(\downarrow\) \\
\midrule
MABR                 & 0.763 & 0.126 & 0.006 & 0.028 \\
- layer[1:5]         & 0.766 & 0.152 & 0.013 & 0.031 \\
- layer[6:9]         & 0.765 & 0.148 & 0.012 & 0.030 \\
- layer[10:12]       & 0.767 & 0.158 & 0.012 & 0.028 \\
\bottomrule
\end{tabular}
\caption{Performance metrics with the bias detector and domain classifier removed at specific layer levels during adversarial training using the MABR method on the sentiment analysis task with Roberta.}
\label{tab:roberta-remove-specific-layers}
\end{table}

Figure \ref{fig:validation_accuracies} and \ref{fig:validation_accuracies-roberta} illustrate the accuracy for each layer of the bias detectors and domain classifiers before and after the adversarial training process for BERT and RoBERTa, respectively. Initially, the accuracy of the bias detectors is notably high. For BERT, all detectors achieve accuracies greater than 0.79 before adversarial training and remain above 0.76 afterward. Similarly, RoBERTa's detectors maintain strong performance, with accuracies exceeding 0.74. This indicates that the bias detectors effectively determine whether the main model succeeds in its task without needing access to the main task labels. This observation supports our assumption that many samples identified by the bias detectors rely on biased features as shortcuts to make predictions, consistent with the findings of \citet{orgad-belinkov-2023-blind}.

Furthermore, we notice that the adversarial training process significantly reduces the accuracy of the bias detectors, demonstrating that the adversarial training has effectively mitigated the bias features in the embedding maps. This makes it harder for the bias detectors to identify easy biases. However, the bias detectors still maintain relatively high accuracy post-training because they are trained during the process simultaneously. As a result, the labels of the samples input to the domain classifier are dynamically refined, which is a significant difference over previous adversarial training methods \citep{elazar-goldberg-2018-adversarial, wang2019balanced, han-etal-2021-diverse}.

We also observe that different layers respond differently to the adversarial training process. As depicted in Figure \ref{fig:validation_accuracies} and \ref{fig:validation_accuracies-roberta}, the early layers behave similarly. The reduction in the accuracy of the bias detector is relatively low, and the accuracy of the domain classifiers remains quite high. This suggests that the lower layers capture fundamental features that are less susceptible to bias, thereby leaving limited room for mitigating bias features without compromising the final accuracy. However, this does not imply that mitigation at the lower levels is unimportant. As evidenced by the data in Tables \ref{tab:bert-remove-specific-layers} and \ref{tab:roberta-remove-specific-layers}, if we remove the adversarial training process from the lower layers (layer 1 to 5), the fairness metrics still degrade significantly. 

\section{Conclusion}

In this paper, we introduced MABR, a novel adversarial training strategy that mitigates biases across various encoder layers of LLMs. By employing multiple auxiliary classifiers to capture different aspects and levels of bias, our approach effectively identifies and reduces social biases without prior knowledge of bias types or demographic annotations. This method significantly improves fairness in tasks such as sentiment analysis and occupation classification, matching or exceeding the performance of models requiring detailed demographic insights. Our findings underscore the importance of leveraging the distinct capabilities of different model layers in capturing nuanced bias features, marking a significant advancement in bias mitigation techniques.

\section*{Acknowledgements} 

We appreciate constructive feedback from anonymous reviewers and meta-reviewers. This work is supported by the Natural Sciences and Engineering Research Council of Canada (NSERC), Discovery Grants program.

\bibliography{anthology,custom}

\end{document}